\documentclass[lettersize,journal]{IEEEtran}
\usepackage{amsmath,amsfonts}
\usepackage{algorithmic}
\usepackage{algorithm}
\usepackage{array}
\usepackage[caption=false,font=normalsize,labelfont=sf,textfont=sf]{subfig}
\usepackage{textcomp}
\usepackage{stfloats}
\usepackage{url}
\usepackage{verbatim}
\usepackage{graphicx}
\usepackage{cite}
\usepackage[dvipsnames]{xcolor}

\usepackage[pagebackref,breaklinks,colorlinks]{hyperref}

\usepackage{graphicx}
\usepackage{amsmath}
\usepackage{amssymb}
\usepackage{booktabs}
\usepackage{adjustbox}
\usepackage{comment}
\usepackage[accsupp]{axessibility}  % Improves PDF readability for those with disabilities.
\usepackage{fancyhdr}

\begin{document}

% \title{Accepted to 2023 CVPRW on Event-Based Vision https://tub-rip.github.io/eventvision2023/}

\title{\textcolor{gray}{\fontsize{12}{18}\selectfont{Accepted to 2023 CVPRW on Event-Based Vision @IEEE} \newline \href{https://tub-rip.github.io/eventvision2023/}{https://tub-rip.github.io/eventvision2023/}}\newline \newline \fontsize{24}{18}\selectfont{Asynchronous Events-based Panoptic Segmentation using Graph Mixer Neural Network}}

\author{Sanket Kachole$^{1}$ \hspace{0.5cm} Yusra Alkendi$^{2,3}$ \hspace{0.5cm} Fariborz Baghaei Naeini$^{1,4}$  \hspace{0.5cm} Dimitrios Makris$^{1}$   \hspace{0.5cm} Yahya Zweiri$^{2,3}$
\\
Department of Computer Science, Kingston University, London, UK$^1$  %\hspace{0.5cm} Ipsotek, an Eviden Company, London $^3$ \hspace{0.5cm} 
\\Advanced Research
and Innovation Center (ARIC), Khalifa University, Abu Dhabi, UAE$^2$ 
\\
Department of Aerospace Engineering, Khalifa University of Science and
Technology, Abu Dhabi, UAE$^3$
\\Ipsotek, an Eviden Company, London$^4$
\\
{\tt\small \{K1742163,f.baghaeinaeini,d.makris\}@kingston.ac.uk$^1$ \hspace{0.25cm} \{yusra.alkendi,yahya.zweiri\}@ku.ac.ae$^2$} }

% The paper headers
%\markboth{Journal of \LaTeX\ Class Files,~Vol.~14, No.~8, August~2021}%
% \fontsize{12}{18}\selectfont\markboth{{A\MakeLowercase{ccepted to} 2023 CVPRW \MakeLowercase{on} E\MakeLowercase{vent} B\MakeLowercase{ased} V\MakeLowercase{ision}}  \href{https://tub-rip.github.io/eventvision2023/}{\MakeLowercase{https://tub-rip.github.io/{eventvision2023}/}}}%
% {Shell \MakeLowercase{\textit{et al.}}: A Sample Article Using IEEEtran.cls for IEEE Journals}

% \chead{\textbf{Accepted to 2023 CVPRW on Event-Based Vision https://tub-rip.github.io/eventvision2023/}}
%\chead{\href{https://github.com/sanket0707/GNN-Mixer.git}{https://github.com/sanket0707/GNN-Mixer.git}}

%\IEEEpubid{0000--0000/00\$00.00~\copyright~2021 IEEE}
%Remember, if you use this you must call \IEEEpubidadjcol in the second
% column for its text to clear the IEEEpubid mark.

\maketitle

\begin{abstract}

In the context of robotic grasping, object segmentation encounters several difficulties when faced with dynamic conditions such as real-time operation, occlusion, low lighting, motion blur, and object size variability. In response to these challenges, we propose the Graph Mixer Neural Network that includes a novel collaborative contextual mixing layer, applied to 3D event graphs formed on asynchronous events. The proposed layer is designed to spread spatiotemporal correlation within an event graph at four nearest neighbor levels parallelly. We evaluate the effectiveness of our proposed method on the Event-based Segmentation (ESD) Dataset, which includes five unique image degradation challenges, including occlusion, blur, brightness, trajectory, scale variance, and segmentation of known and unknown objects. The results show that our proposed approach outperforms state-of-the-art methods in terms of mean intersection over the union and pixel accuracy. Code available at: \href{https://github.com/sanket0707/GNN-Mixer.git}{https://github.com/sanket0707/GNN-Mixer.git}

\end{abstract}


\section{Introduction}

 %   1. General Context (importance of Robotic grasping, applications, need for segmentation), 
Object grasping is a crucial task for robots with applications in manufacturing, logistics, healthcare, and household tasks \cite{sanket20163Tracker,Garg2020SemanticsSurvey}. However, detecting and segmenting objects accurately in the robot's environment is challenging due to occlusions, complex geometries, and dynamic backgrounds \cite{Dinakaran2023AGrasping}. Panoptic segmentation aims to simultaneously segment foreground objects and background regions in an image. Integrating panoptic segmentation into object grasping enables robots to perceive their environment better and perform more complex tasks efficiently.
    
 %   2. Challenges (previous work in segmentation, challenges related to robotic grasping), 
 
Common challenges in panoptic segmentation are due to cluttered scenes, object geometry and appearance variability \cite{Marcuzzi2023Mask-BasedDriving,Wang2022SFNet-N:Scenes}, occlusions, motion blur \cite{Mohan2022AmodalSegmentation} and low temporal resolution \cite{Montano2016NetworkPaper} in traditional cameras. High latency can cause delays in processing sensor data, resulting in slower response times and reduced accuracy in performing tasks. Recent progress in object segmentation using State-of-the-Art Graph Neural Networks with a transformer mechanism \cite{AlkendiNeuromorphicNetwork,Alkendi2022NeuromorphicTransformers} introduces additional constraints, as both panoptic segmentation and grasp planning must be performed quickly and efficiently. To address these challenges, more sophisticated algorithms and techniques are needed to better handle the variability and complexity of real-world environments. % Relevant research studies are cited in support of these claims.

\begin{figure}[!]{
\centering
\setlength{\fboxrule}{1.pt}%
\centering
\begin{adjustbox}{width=0.48\textwidth}
  \centering
  {\Large
\begin{tabular}{ccc}

 \fbox{\includegraphics[width=\columnwidth,height=7.1cm]{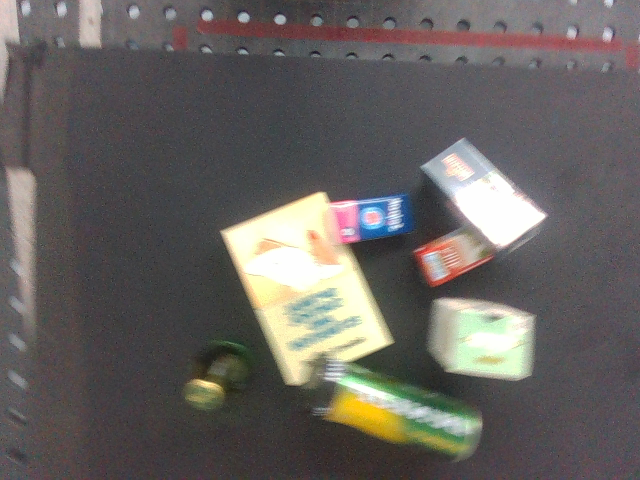}}\centering
&\fbox{\includegraphics[width=\columnwidth,height=7.1cm]{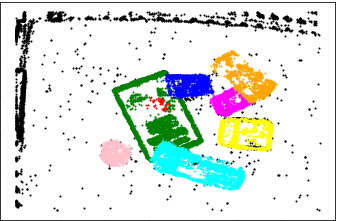}}\centering
&\fbox{\includegraphics[width=\columnwidth,height=7.1cm]{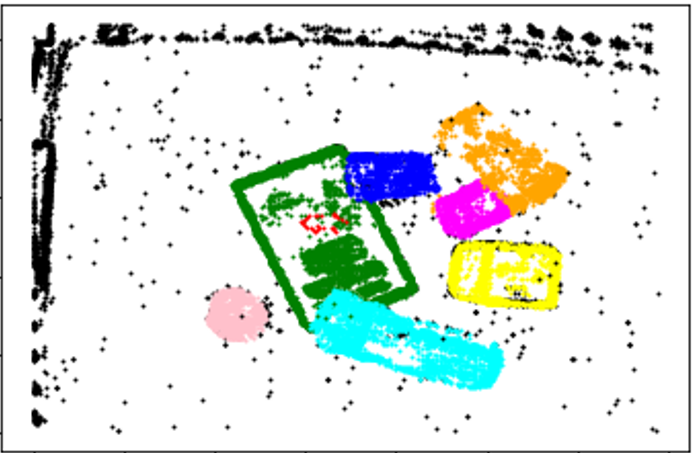}}

\\
\Huge{\begin{tabular}{c} (A)\end{tabular}} &\Huge{(B)}& \Huge{(C)} \\
% &\Huge{SpikeMS\cite{SpikeMS}}\\ 
\end{tabular}}
\end{adjustbox}
 \setlength{\belowcaptionskip}{-12pt} 
\vspace{-10pt}
\caption{Panoptic segmentation results of the proposed learning-based algorithm (GMNN) applied on the ESD dataset\cite{Huang2023AEnvironment}. (A) APS image for visualization only. (B) Ground truth approximate events (C) Segmented events using the GMNN algorithm.}
\label{Sample_Qualitative}}
\end{figure}

We propose the Graph Mixer Neural Network (GMNN) for event-based panoptic segmentation. Our proposed model maintains the asynchronous nature of event streams and leverages spatiotemporal correlations to infer the scene. The key technical contribution is the novel Collaborative Contextual Mixing (CCM) layer within a graph neural network architecture that enables the parallel mixing of event features generated from multiple sets of neighborhood events. Our proposed model achieves state-of-the-art performance on the ESD dataset  \cite{Huang2023AEnvironment} which consists of robotic grasping scenes captured using an event camera mounted next to the gripper of a robotic arm. The dataset includes scenes with variations in object clutter size, arm speed, motion direction, distance between the object and camera, and lighting conditions. Specifically, it achieves superior results in terms of mean Intersection Over Union (mIoU) and pixel accuracy, while also demonstrating significant improvements in computational efficiency compared to existing state-of-the-art methods.
Fig. \ref{Sample_Qualitative} shows segmentation results obtained when testing our GMNN on a sample from the ESD dataset.

%    5. Paper outline.

Previous work is discussed in Section II. Our proposed architecture is described in detail in Section III. The validation of our method through experimental results and an ablation study are presented in Section IV. Finally, the conclusion and scope for further research are outlined in Section V.

%%%%%%%%%%%%%%%%%%%%%%%%%%%%%%%%%%%%%%%%% Literature Review %%%%%%%%%%%%%%%%%%%%%%%%%%%%%%%%%%%%%%%%%%%

\section{Related Work}
\subsection{\textbf{Image segmentation Methods}}

Thresholding algorithms have fixed thresholds \cite{N2016ImageImages}  and lack contextual information, while clustering techniques  \cite{Thilagamani2011ObjectClustering,AlkendiNeuromorphicNetwork} can adapt to variable structures but are sensitive to initial conditions and may lead to over-segmentation. Deep learning methods
\cite{Ronneberger2015U-net:Segmentation},\cite{ Chen2018DeepLab:CRFs},\cite{Long2015FullySegmentation}  produce dense predictions but may ignore small objects and details. Event-based methods have advantages as they can handle motion blur and high dynamic range \cite{AlkendiNeuromorphicNetwork} but may require labeled images such as the Event-based Semantic Segmentation (ESS) method \cite{Sun2022ESS:Images}  and have limitations in segmenting small objects \cite{Alonso2019EV-SegNet:Cameras}. Multiple modalities can be integrated to leverage complementarity, with CMX using transformer-based architecture \cite{Liu2023CMX:Transformers} and Bimodal SegNet \cite{Kachole2023BimodalGrasping} fusing RGB with event frames. Although these methods demonstrate promising results, their limitations include overlooking the high temporal resolution of event-based data. % or difficulties in edges, and object contour features.

\subsection{\textbf{Graph neural network methods}}

The adoption and evolution of GNNs in computer vision applications has been remarkable in recent years\cite{Chen2022APerspective}. Asynchronous Event-based GNNs (AEGNNs) \cite{Schaefer2022AEGNN:Networks} extend GNNs to process events as evolving spatiotemporal graphs. However, using conventional deep neural networks to process dense representations of events eliminates their sparsity and asynchronous nature, leading to computational and latency constraints. GNN-transformer \cite{AlkendiNeuromorphicNetwork} addresses the problem by utilizing spatiotemporally evolving graphs that can be efficiently and asynchronously processed using GNNs. TactileSGNet \cite{Gu2020TactileSGNet:Recognition} utilized a spiking-GNN \cite{Schliebs2013EvolvingSurvey} and a GNN-Transformer algorithm to perform event-based recognition of tactile objects and classify active event pixels by leveraging the EventConv message-passing framework to capture spatiotemporal correlations among events while preserving their asynchronous nature \cite{Alkendi2022NeuromorphicTransformers}.

% \begin{comment}Several architecture designs have been proposed to enhance the depth and capacity of asynchronous graph neural networks while ensuring efficient processing. These designs incorporate node update pruning, early temporal aggregation via max pooling, novel efficient Look-Up-Spline Convolutions, and directed event graph processing in early layers. Nonetheless, the approach would have benefitted from a recurrence mechanism as it can mitigate problems associated with a limited number of events. \cite{Gehrig2022PushingCameras}
% TactileSGNet employed a spiking graph neural network to perform event-based recognition of tactile objects. To leverage the local connectivity of voxels, several techniques were introduced to organize tactile data into a graph structure \cite{Gu2020TactileSGNet:Recognition}. The GNN-Transformer architecture was utilized to classify active event pixels in the raw stream as either real log-intensity variation or noise. This was achieved by leveraging the EventConv message-passing framework, which can capture spatiotemporal correlations among events while still preserving their asynchronous nature \cite{Alkendi2022NeuromorphicTransformers}. Researchers employed a spatiotemporal graph-based approach to address two challenges: event-cluster assignment and motion model fitting, both of which were tackled iteratively. The proposed method offers several benefits, including globally consistent and locally coherent results \cite{Zhou2021Event-BasedCuts}.
%  \end{comment}

Graph Transformer Neural Network (GTNN) applies a self-attention mechanism to motion segmentation in asynchronous event-based vision data streams using 3D graphs \cite{AlkendiNeuromorphicNetwork}. However, long-range dependencies pose a challenge for transformer-based models due to their spatially variant nature. Event-based Transformer (EvT) resolves this by creating event frames and employing sparse patch-based event data representation with attention mechanisms \cite{Sabater2022EventProcessing}. Despite the promise of transformer-based methods, modeling long-range dependencies is difficult due to their spatially variant nature. 
MLP-like architectures applied in 3-D point clouds outperform transformers and CNNs in handling position-sensitive information using simple token and channel-mixing MLPs \cite{Zhao2020PointTransformer}, without self-attention mechanisms, on large-scale data. However, Metaformer shows that general architecture formulation is more critical than specific interaction strategies, achieving remarkable results by replacing token-mixing with average pooling \cite{Yu2022MetaFormerVision}. The PointMixer method improves feature mixing within and between point sets using a universal point set operator instead of token-mixing MLPs, resulting in better parameter efficiency and accuracy with Softmax substitution \cite{Choe2022PointMixer:Understanding}.

MLPs excel in different applications but not in asynchronous event-based vision tasks, while GNNs have yet to explore modern MLP-like techniques \cite{Zhao2020PointTransformer, Choe2022PointMixer:Understanding}. The current use of K-Nearest Neighbors (KNN) is insufficient to improve feature mixing within GNNs, and novel approaches are required for challenging tasks such as event-based panoptic segmentation.

%###################################### Methodology ################################
\section{Methodology}

\subsection{Prerequisite}

\subsubsection{\textbf{Event-based vision data}}
%-\newline
Event-based vision cameras respond to changes in log intensities by capturing pixel-level changes called events. A continuous stream of events is mathematically represented by a sequence of tuples comprising $i$th event location $(x_i, y_i)$, timestamp $t_i$, and polarity $z_i$ \cite{Naeini2020Dynamic-vision-basedNetworks,Naeini2022EventMeasurements}: 
    
\begin{equation}
    {(x_1, y_1, t_1, z_1), (x_2, y_2, t_2, z_2), ..., (x_n, y_n, t_n, z_n)}
\end{equation}

\subsubsection{\textbf{Graph Neural Network}}
%-\newline
Graph Neural Networks (GNNs) consist of nodes or vertices $V$, connected by edges or links $L$ \cite{Gao2019GraphU-Nets}. Mathematically, a GNN can be expressed as 
%\begin{equation}
    $G=(V,L)$.
%\end{equation}
Each node $i \in V$ takes as input the weighted sum of the output values $q_i$ of its incoming edges  $o_i$, and produces an output value %$y_i$.
%\begin{equation}
    $r_i = f_i(o_i^T q_i)$.
%\end{equation}

\subsubsection{\textbf{Multi-Layer Perceptron (MLP)}}
%-\newline
Each node $i \in V$ assumes an initial feature vector $p_i^0$ at layer $0$, which is transformed using an MLP to produce new feature vectors $p_i^l$ at each MLP layer $l$:
\begin{equation}
    p_i^{l} = \sigma \Bigg(W^l \sum_{j \in N_i} \frac{1}{c_{i,j}} p_j^{l-1} + b^l \Bigg)
\end{equation}
where $\sigma$ is a non-linear activation function, $N_i$ denotes the set of neighboring nodes of queried node, $W^l$ is weight matrix, $b^l$ is bias vector of the MLP at layer $l$ and $c_{i,j}$ is a normalization factor \cite{Lv2022CM-MLP:Image,Choe2022PointMixer:Understanding}.

\subsubsection{\textbf{K-Nearest Neighbors on Graph Nodes}}
%-\newline
The k-Nearest Neighbor (kNN) method is commonly used to process event data locally by considering proximity, resulting in an index map of neighboring nodes represented as $M_i$ \cite{Gao2009VisibleDatabases}. Let's assume a set of nodes $N = \{n_i\}_{i=1}^N$ with corresponding features $P = \{p_i\}_{i=1}^N$. %, where, $n_i$ is the position of $ith$ node and $p_i$ are corresponding features.
For a query node $n_i$, an index map $M_{k,i}$ of the $k$ closest nodes can be calculated using kNN as: 
\begin{equation}
     M_{k,i} = kNN(N,k,n_i)
\end{equation}

The corresponding feature set for this kNN is $P_{k,i}=\{p_j \in P | j \in M_{k,i}\}$.

\begin{figure*}[t!]
\resizebox{\hsize}{!}{\includegraphics[clip=true width=0.8\linewidth]{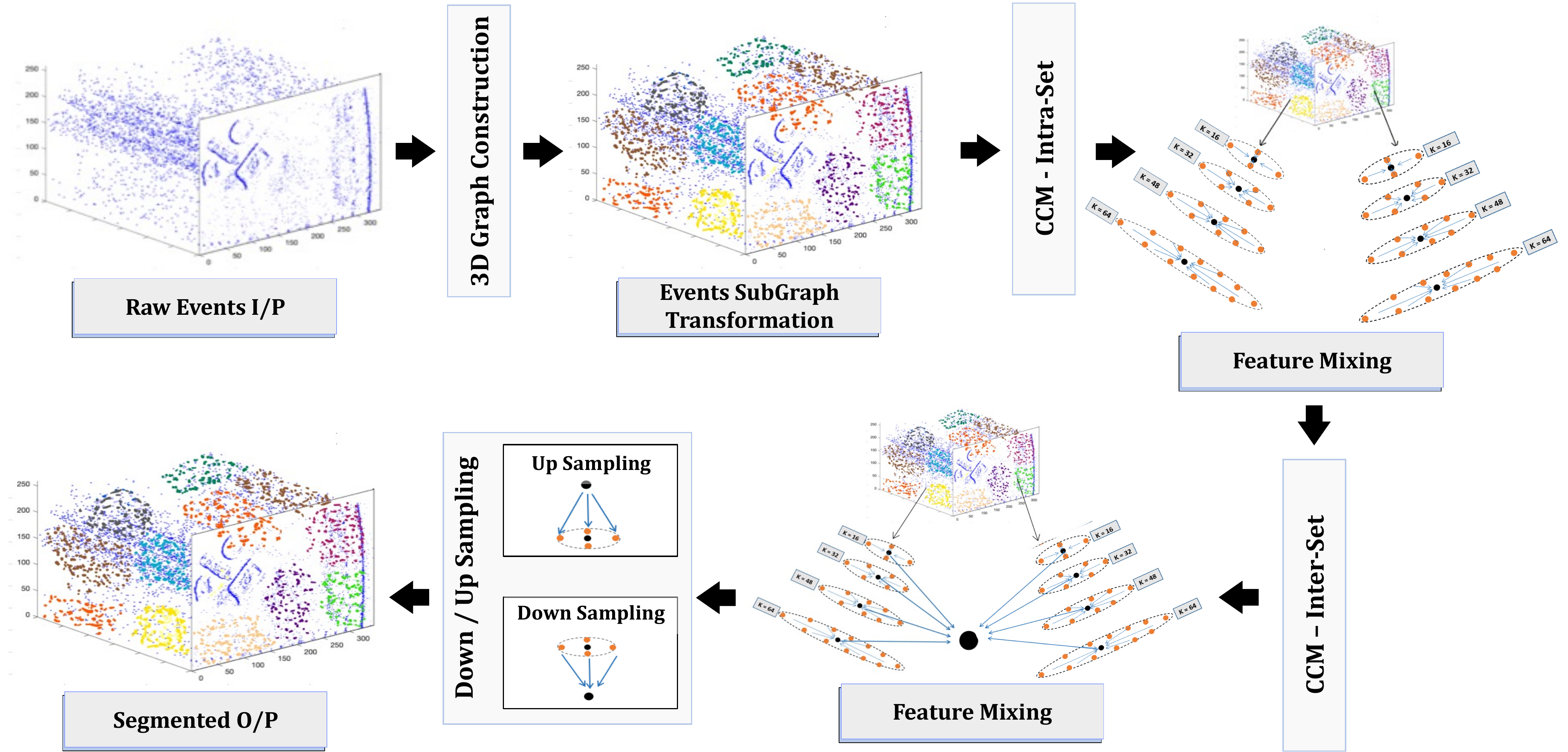}}
\caption{\footnotesize
{ Proposed Framework - Graph Mixer Neural Network (GMNN) for panoptic segmentation of asynchronous event data in a robotic environment. GMNN operates on a 3D- graph constructed of DVS events acquired within a temporal window, encapsulating its spatiotemporal properties. Subgraphs of spatiotemporally neighboring events are then constructed (colored event in step 2) where each is processed by various nonlinear operations within Mixer and sampling modules to perform segmentation. }
}
\label{fig:GMNN Famework}
\setlength{\belowcaptionskip}{-18pt} 
\end{figure*}

\begin{figure}[h]
      \centering
\includegraphics[width=0.8\linewidth]{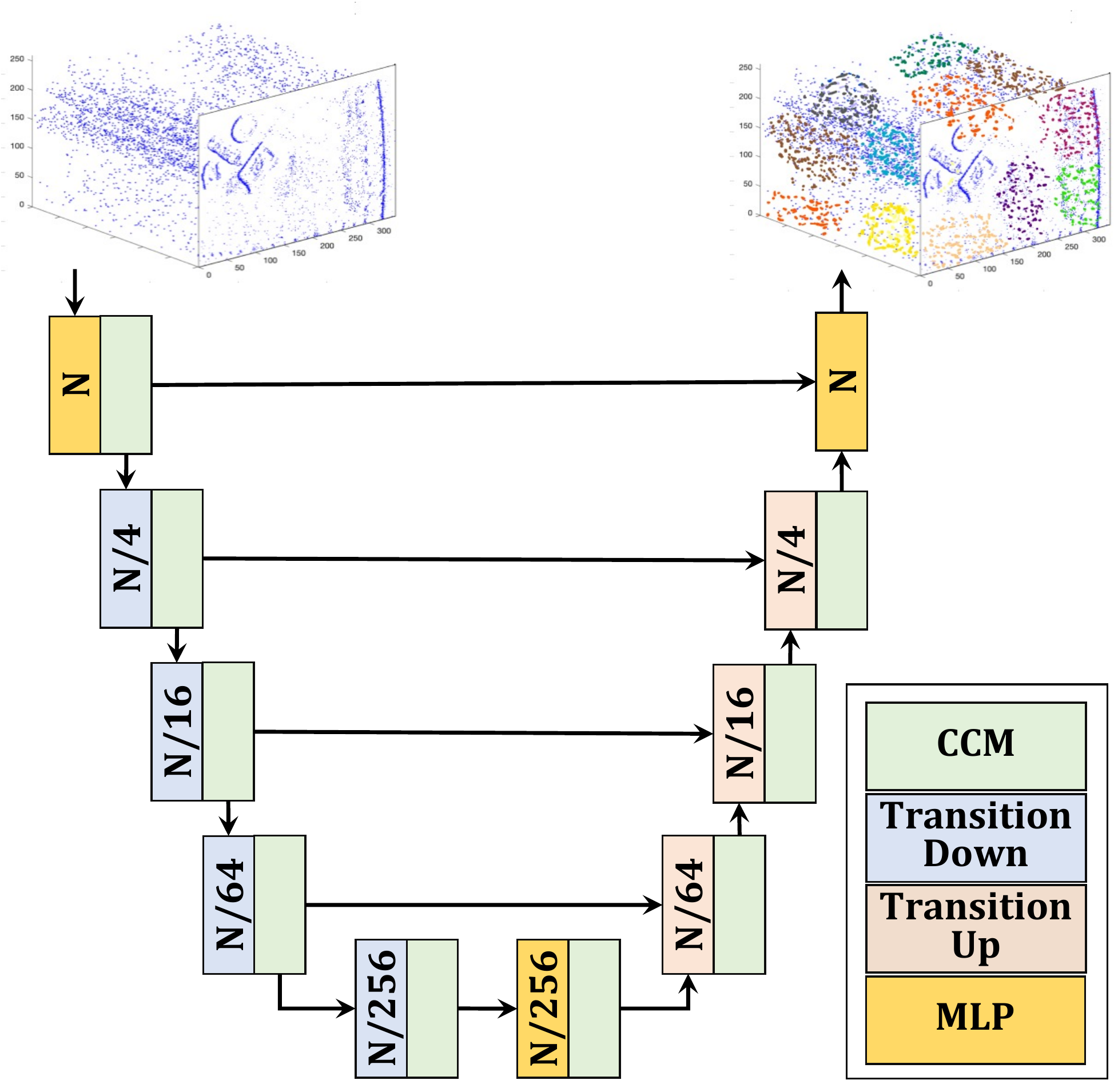}
 \setlength{\belowcaptionskip}{-12pt} 

\caption{Graph Mixer Neural Network (GMNN). Note that "\textbf{N}" represents the number of nodes (i.e., Events) per graph.}

\label{fig:Architecture}
\end{figure}

\subsection{\textbf{Graph Mixer Neural Network }}
In this section, the framework of the Graph Mixer Neural Network, depicted in Fig. \ref{fig:GMNN Famework}, is explained in detail. The first section \ref{section: 3-D Graph Construction} describes a method for constructing a 3D graph to represent a series of events that occur within a pre-determined time interval. The graph is constructed based on the most recent $N_{max}$ events, and a k-Nearest Neighbor (kNN) search connects each node with its k-nearest neighboring nodes. The next section \ref{section: Collaborative contextual mixing}  the Collaborative Contextual Mixing (CCM) method, a novel approach for disseminating event features across various sets, is described. Further, in section \ref{section:Transition down block} a transition down block to down-samples the graph, and in section \ref{section:Transition up block} a transition up the block to upsample graph nodes are discussed.

\subsubsection{\textbf{3-D Graph Construction}}
%-\newline
\label{section: 3-D Graph Construction}

A 3D graph $G$ represents a series of events that occur within a pre-determined time interval $T$. Each node $i$ represents an event in the stream, with features $e_i=[x_i,y_i,t_i]$. Event polarityvaries with camera parameter settings, and therefore limits the generalizability of the proposed algorithm. Therefore, we excluded polarity from the graph node features, following \cite{Choe2022PointMixer:Understanding}.

The number of events triggered by changes in pixel intensities in the output event stream depends on the camera's speed. Higher speed results in more events, while lower speed results in fewer events. These conditions make it difficult to determine spatiotemporal event relationships due to redundant data and high computational demands. In addition, a memory limit imposes a maximum number of events that can be stored, causing the potential loss of important information from earlier events. It also affects the ability to determine spatiotemporal relationships, resulting in reduced accuracy and precision. Therefore, the choice of $N_{max}$ needs careful consideration to balance memory usage and accuracy. To address these issues and minimize memory consumption, the graph is constructed within each time interval T with a given maximum number of nodes  $N_{max}$. If the number of events within this time window exceeds the $N_{max}$, e.g. because of the scene and camera dynamics, only the most recent $N_{max}$ events are preserved. %, which vary based on the scene and camera dynamics.
The method can be applied across different domains, regardless of the number of events triggered within the temporal window, thanks to the feature of graph-based neural networks to operate on graphs of varying sizes.

Spatiotemporal distances are scaled by dividing the spatial distances by the maximum spatial distances X and Y and the temporal distances by the maximum temporal distance T. This normalization ensures that the spatial and temporal distances are on the same scale and have equal weight in the calculation of the spatiotemporal distances. A k-Nearest Neighbor (kNN) search connects each node $i$ with its corresponding feature $p_i$  with its $k$-nearest neighboring nodes $j$ and their corresponding  features $p_j$ using 3D normalized spatiotemporal distances. The resulting spatiotemporal neighborhoods are called sub-graphs, each with $k+1$ nodes. Once all subgraphs are constructed, each will pass through the nonlinear operations of the Mixer Layer where each node (event) features are encoded, and the Sampling Up/Down where graph nodes are convolved/deconvolved.

\subsubsection{\textbf{Collaborative contextual mixing}}
%-\newline
\label{section: Collaborative contextual mixing}
Feature mixing is one of the important aspects in graph neural networks to understand the relationships between nodes. Existing feature mixing methods employ kNN-based subsampling to disseminate the features. We argue that the use of k-nearest neighbors (kNN) solely is inadequate as it confines an event to collect information from a restricted neighborhood. Considering the sparse and asynchronous nature of events, the event-based vision domain demands advanced techniques. 

\begin{figure}[h]
      \centering
\includegraphics[width=0.75\linewidth]{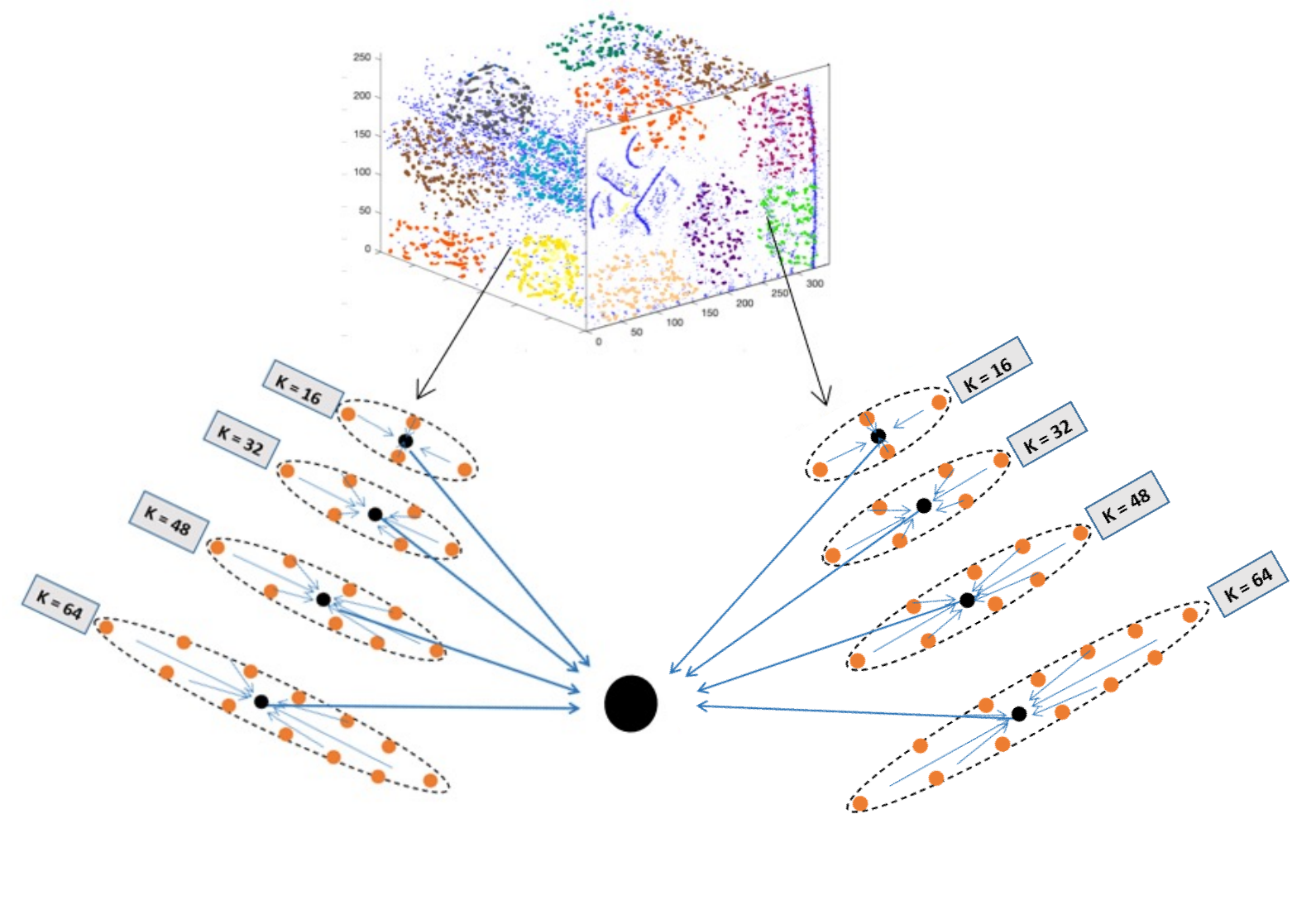}
 \setlength{\belowcaptionskip}{-12pt} 
\caption{Collaborative Contextual Mixing depicting CCM intra-set and Inter-Set feature mixing.}
\label{fig:CCM}
\end{figure}

We introduce the Collaborative Contextual Mixing (CCM) method as a novel approach to disseminating event features across various sets. Thus, distributing the event features among multiple levels of the nearest neighbors in parallel and then aggregating them using a weighted sum results in a more effective feature mixing as shown in Fig. \ref{fig:CCM}.  

First, a spatial pyramidal block of $kNNs$ is applied simultaneously at four levels with $k\in\{16,32,48,64\}$ ($kNN_1$ to $kNN_4$) which produces four index maps $M_{k,i}$, each with a corresponding feature set $P_{k,i}$. This choice was made based on the network, in accordance with the ablation study conducted on its hyperparameters and variants.
For each query node $n_i$, at level k, a score vector $s = [s_1,.....s_k]$ is computed:

\begin{equation}
    s_j = g_2([g_1(x_j);\delta(e_i - e_j)],  where j \in M_{k,i}
    \label{eq:Scoring}
\end{equation}

where $g$ is a channel mixing MLP, $\delta$ refers to the relative positional encoding MLPs and $p_j$  is a $j$-th element of the feature vector set $P_i$. The computed score vector is then passed into the following function to compute the output feature vector $u_{k,i}$ as follows:

\begin{equation}
    u_{k,i} = {\sum_{j \in M_{k,i}}} softmax(s_j) * {g_3} (x_j)
    \label{eq:Compute Feature Vector}
\end{equation}

where the softmax function normalizes the spatial dimension. The symbol `$*$' indicates element-wise multiplication. Let $u_{k,i}$ denote the new feature vector obtained after aggregating $k$ adjacent nodes. %It is applied four times on the 16, 34, 48, and 64 nearest neighbors, resulting in four output feature vectors from $KNN1$ to $KNN4$, respectively. 
As the KNN is applied 4 times it will produce output feature vectors for each level, namely $u_{1,i}$, $u_{2,i}$, $u_{3,i}$, and $u_{4,i}$ and are subsequently aggregated using weighted sum as follows to obtain the final output feature vector $u_i$:

\begin{equation}
u_i = w_1 u_{1,i} + w_2 u_{2,i} + w_3 u_{3,i} + w_4 u_{4,i}
\label{eq: final output feature vector}
\end{equation}

$w_1$, $w_2$, $w_3$, and $w_4$ represent the weights assigned to each output feature vector. Followed by intra-set mixing, to mix the information between the sets the inter-set mixing method is applied. This can be understood as the inverse of the kNN, and  the inverse index mapping $M_{k,i}^{-1}$ is defined (see Eq.\ref{eq:inverse map}, which  finds the set of indices $j$ that includes event $e_i$. In this manner event features from the neighboring event sets are mixed \cite{Choe2022PointMixer:Understanding}.

%The $w_1$, $w_2$, $w_3$, and $w_4$ represent the weights of each output feature vector. $0.10$, $0.20$, $0.30$, and $0.40$     

\subsubsection{\textbf{Transition down block}}
%-\newline
\label{section:Transition down block}
The first step in the transition down block of the proposed method is to perform sampling of graph nodes using the farthest point sampling algorithm. The resulting sample nodes, denoted as $G_s$, are a subset of the original graph nodes $G_o$, i.e., $G_s \subset G_o$. The downsampled nodes are then used to compute their neighbors in the original graph using the kNN algorithm at four different levels k, which produces index maps denoted as $M_k$. By applying Eq.\ref{eq:Scoring} and Eq.\ref{eq:Compute Feature Vector} with the calculated index mapping, the features of the original graph nodes are passed to the sample graph nodes.

For an original graph $G$ with $E_i$ nodes denoted as $G(E_i)$, the kNN algorithm is used to downsample it to $G(e_i)$. Mathematically,

\begin{equation}
    M_{k,i}^s = KNN(E_i, k, e_i)
\end{equation}

The kNN algorithm acts as a reduction factor in the transition down block, reducing the cardinality of the 3D graph and enabling the convolution of graph nodes. Specifically, if the original graph $G$ has $N$ nodes and a requested reduction factor of 4, the transition down module produces a new graph with $N/4$ nodes. %In this study, the value of $k$ is chosen to be 16 empirically.

\subsubsection{\textbf{Transition up block}}
%-\newline
\label{section:Transition up block}
 
The transition-up block samples graph nodes from the transition-down and original graph node sets, without using the kNN function due to asymmetric neighbors. It utilizes the index mapping computed in the transition-down block to apply inverse mapping and upsample the nodes as:  
\begin{equation}
\label{eq:inverse map}
    M_{k,i}^{-1} = \{j|i \in M_j\}
\end{equation}
The resulting mapping is used to calculate the original sample using Eq.\ref{eq:Scoring} and Eq.\ref{eq:Compute Feature Vector}, as shown in the Fig. \ref{fig:GMNN Famework}. The transition-up module maps features from reduced graph dataset, G'(p2), to its superset, G'(p1) (where G'(p1) $\supset$ G'(p2)), without requiring an additional kNN search. Concatenating interpolated features of G'(p1) with the corresponding encoder stage's features via a skip connection enhances the features learned at the same level of the transition down block.

\subsection{\textbf{Proposed Network Architecture}}
\label{section: Proposed Network Architecture}

The proposed GMNN architecture (Fig. \ref{fig:Architecture}) has four components: MLP blocks, transition down, transition up, and Mixer blocks, which form the encoder and decoder. The 3D event graph is passed through the encoder's MLP layer, followed by four downsampling levels that reduce node numbers by a factor of four each. The Mixer block uses the novel CCM method to spread features in parallel. The output of the encoder is then passed into the decoder, which begins with an MLP, followed by four upsampling levels with a factor of four each. Graph nodes are upsampled using the transition up block, and the Mixer block spreads features using the proposed CCM method in \ref{section:Transition up block}. The header block includes MLPs without a pooling layer for dense prediction tasks. The architecture adopts a deep pyramid-style structure to obtain global features by progressively downsampling nodes.
In contrast to the conventional Graph Neural Network (GNN) approach, where graphs are constructed prior to network nonlinear operations and predictions, the Graph Mixture Neural Network (GMNN) employs a novel strategy of constructing subgraphs from each input graph and processing them in parallel within the Mixer layer and sampling modules. This approach facilitates the identification of spatiotemporal correlations between events and effectively captures the motion dynamics.

%###################################  Experiments 

\section{Experiments}

\subsection{\textbf{Dataset} }

The ESD dataset \cite{Huang2023AEnvironment} includes 17,186 annotated images and 177 labeled event streams, captured using a Davies346 sensor mounted at end of the robotic arm. It has variations in camera motion, arm speed, lighting conditions, and cluttered scenes. The dataset has instance-wise annotations for 15 object classes grouped into 6 categories. The training set (ESD-1) consists of 13,984 images of 10 known objects, while the testing set (ESD-2) consists of 3,202 images of 5 unknown objects that are not in ESD-1. Please refer to \cite{Huang2023AEnvironment} for a detailed explanation of the experimental setup.

\subsection{\textbf{Training }}

In this study, in order to enable a fair comparison with state-of-the-art (SOTA) methods, we adopt an \textit{effective} training scheme proposed by \cite{AlkendiNeuromorphicNetwork}, that takes advantage of large amounts of training data while reducing computational requirements. The proposed scheme involves dividing the full training dataset into L subsets and exposing the neural network to only one of the subsets during each iteration, while the remaining subsets remain inactive. The network trains on a specific subset once every L iteration, resulting in complete training on the entire dataset after L epochs. This differs from conventional training methods where the entire dataset is used to update the neural network weights in every epoch.  During the training process, the SGD optimizer is employed with a learning rate of 0.001 to minimize loss. In this study, each 3D event node is assigned to one of ten semantic categories, and the evaluation protocol suggested by Point Transformer is closely followed to ensure fairness. We use the SGD optimizer with a batch size of 4 during training and set the momentum and weight decay values to 0.9 and 0.0001 respectively. The weights $w_1$, $w_2$, $w_3$, and $w_4$  are empirically set to $0.10$, $0.20$, $0.30$, and $0.40$ respectively. The time interval is set to $T=100ms$ and the maximum number of nodes to $N_{max}=10000$, while the maximum spatial distances depend on the resolution of the event camera, therefore $X = 346$ and $Y = 260$.

\subsection{\textbf{Evaluation Metrics}}
%\subsubsection{\textbf{Pixel accuracy}}

Pixel accuracy and mIoU are used to evaluate the performance of panoptic segmentation. Pixel accuracy calculates the percentage of pixels in the image that are classified correctly. To adapt pixel accuracy to event-based vision data, the ratio for each object count of predicted events to ground truth events is calculated. Mean accuracy is then calculated across all objects. This approach provides a way to evaluate object detection models based on event data and accounts for the sparsity of events: 

\begin{equation}
    Acc(d,d') = \frac{1}{N}\sum_{1}^N \frac{d_i}{d'_i}
    \label{eqn: Pixel Accuracy}
\end{equation}

where $d$, $d'$, and $N$ represent the ground truth event set, the predicted event set, and the total number of events respectively.

%\subsubsection{\textbf{Mean Intersection over Union (mIoU)}}

The mIoU, also known as the Jaccard Index, handles better imbalanced binary and multi-class segmentation and is calculated across classes according to Eq.\ref{eqn: mIoU}:

\begin{equation}
    mIoU = \frac{1}{C}\sum_{i}^C \frac {\sum_{i}^N \delta (d_{i,c},1) \delta (d_{i,c},d'_{i,c})}{max (1,\delta (d_{i,c},1) + \delta (d'_{i,c},1) }
    \label{eqn: mIoU}
\end{equation}

\subsection{\textbf{Quantitative Evaluation}}
%\subsubsection{\textbf{Comparison with SOTA approaches}}

In order to assess the efficacy of GMNN for panoptic segmentation, we present an evaluation of each sub-task of the dataset, which includes variations in the number of objects, lighting conditions, motion direction, camera speed, and object size across the entire dataset. Further, a similar evaluation is conducted on the unknown dataset to understand the model accuracy on unknown object segmentation.

\begin{table}[t]
    \centering
    \caption{Segmentation accuracy of known objects in various conditions.}
    \vspace{-10pt}
   \begin{adjustbox}{width=0.48\textwidth}

    \centering
      {\fontsize{10}{12}\selectfont
    % \begin{tabular}{|c|c|c|c|c|c|c|}      
    \begin{tabular}{lccccc}
  
  \specialrule{.15em}{.1em}{.1em}  
        \multicolumn{6}{c}{Exp 1: \textbf{varying clutter objects}, Bright light, 62cm height,  Rotational motion, 0.15 m/s speed } \\
  \specialrule{.1em}{.1em}{.1em} 
        Method & 2 Obj & 4 Obj & 6 Obj & 8 Obj & 10 Obj   \\  %& \begin{tabular}{c} Overlapped \\ Events \\ (\% of FP)\end{tabular}
  \specialrule{.1em}{.1em}{.1em}   
  
        EV-SegNet\cite{Alonso2019EV-SegNet:Cameras} & 82\% & 73\%  & 67\%  & 54\% & 51\%     \\
        ESS\cite{Sun2022ESS:Images} & 86\%  & 76\% &   68\% &  64\%  &  60\% \\
        GTNN\cite{AlkendiNeuromorphicNetwork} & 89\%  & 86\%  & 84\%  &  77\%  &  71\% \\
        GMNN (ours) & \textbf{97\%}  & \textbf{96\%} & \textbf{91\%} &  \textbf{89\%} &  \textbf{87\%} \\
   \specialrule{.15em}{.1em}{.1em}

   \specialrule{.15em}{.1em}{.1em}  
        \multicolumn{6}{c}{Exp 2: 6 Objects, \textbf{varying lighting conditions}, 62cm height, Rotational Motion, 0.15 m/s speed.} \\
  \specialrule{.1em}{.1em}{.1em} 
        Method & Bright Light & Low light &  &  &    \\  %& \begin{tabular}{c} Overlapped \\ Events \\ (\% of FP)\end{tabular}

        EV-SegNet\cite{Alonso2019EV-SegNet:Cameras} & 76\% & 75\%   &  &  &      \\
        ESS\cite{Sun2022ESS:Images} & 79\%  & 78\%  &    &   &   \\
        GTNN\cite{AlkendiNeuromorphicNetwork} & 81\% & 79\%  &   &    &   \\
        GMNN (ours) & \textbf{95\%}  & \textbf{94\%} &  &   &   \\  
   \specialrule{.15em}{.1em}{.1em}

   \specialrule{.15em}{.1em}{.1em}  
        \multicolumn{6}{c}{Exp 3: 6 Objects, Bright Light, 62cm height, \textbf{Varying directions of motion}, 0.15 m/s speed.} \\
  \specialrule{.1em}{.1em}{.1em} 
        Method & Linear & Rotational & Partial Rotational &  &    \\  %& \begin{tabular}{c} Overlapped \\ Events \\ (\% of FP)\end{tabular}

        EV-SegNet\cite{Alonso2019EV-SegNet:Cameras} & 65\% & 73\%  & 69\%  &  &      \\
        ESS\cite{Sun2022ESS:Images} & 68\%  &  78\% &  74\%   &   &   \\
        GTNN\cite{AlkendiNeuromorphicNetwork} & 75\%  &  89\%  & 78\%   &    &   \\
        GMNN (ours) & \textbf{84\%}  &  \textbf{93\%}  &  \textbf{90\%}  &   &   \\  
   \specialrule{.15em}{.1em}{.1em}

   \specialrule{.15em}{.1em}{.1em}  
        \multicolumn{6}{c}{Exp 4: 6 Objects, Bright Light, 62cm height, Rotational motion, \textbf{Varying speed}.} \\
  \specialrule{.1em}{.1em}{.1em} 
        Method & 0.15 m/s & 0.3 m/s &   0.1 m/s &  &    \\  %& \begin{tabular}{c} Overlapped \\ Events \\ (\% of FP)\end{tabular}
  \specialrule{.1em}{.1em}{.1em}   
  
        EV-SegNet\cite{Alonso2019EV-SegNet:Cameras} & 69\%  & 60\%  &  56\%  &  &      \\
        ESS\cite{Sun2022ESS:Images} & 72\%  &  63\%  &  59\%   &   &   \\
        GTNN\cite{AlkendiNeuromorphicNetwork} &  75\%  &  71\%  &  63\%    &    &   \\
        GMNN (ours) & \textbf{93\% }  &  \textbf{91\%}  &  \textbf{87\%}  &   &   \\  
   \specialrule{.15em}{.1em}{.1em}

   \specialrule{.15em}{.1em}{.1em}  
        \multicolumn{6}{c}{Exp 5: 6 Objects, Bright Light, \textbf{Varying camera height}, Rotational motion, Varying speed.} \\
  \specialrule{.1em}{.1em}{.1em} 
        Method & 62 cm & 82 cm &    &   &    \\  %& \begin{tabular}{c} Overlapped \\ Events \\ (\% of FP)\end{tabular}
  \specialrule{.1em}{.1em}{.1em}   
  
        EV-SegNet\cite{Alonso2019EV-SegNet:Cameras} & 76\%  &  74\%  &   &  &      \\
        ESS\cite{Sun2022ESS:Images}& 82\%  & 75\%  &     &   &   \\
        GTNN\cite{AlkendiNeuromorphicNetwork} &  85\% &   83\%  &     &    &   \\
        GMNN (ours) & \textbf{97\% }  &  \textbf{93\%}  &    &   &   \\  
   \specialrule{.15em}{.1em}{.1em} 
   \specialrule{.15em}{.1em}{.1em}

    \end{tabular}}
    \end{adjustbox}
    \label{tab: Segmentation Accuracy in Various conditions}
     \setlength{\belowcaptionskip}{-12pt} 
\end{table}

The first experiment used a subset of the testing dataset with varying clutter levels of 2, 4, 6, 8, and 10 objects. More objects meant more occlusions and a more challenging scenario, evident in the segmentation accuracy results shown in experiment 1 of the table, \ref{tab: Segmentation Accuracy in Various conditions}. Accuracy scores for state-of-the-art models decreased from 82\%-89\% for 2 objects to 51\%-70\% for 10 objects, a reduction of 31\%-19\%. In contrast, the proposed GMNN model dropped only by 10\% and outperformed the other methods for any number of scene objects.

State-of-the-art methods in comparison to GMNN show relatively poor segmentation accuracy in both bright lighting conditions, as shown in experiment 2 of table \ref{tab: Segmentation Accuracy in Various conditions}: 76$\%$, 80$\%$ and 81$\%$ for EV-SegNet, ESS and GTNN respectively. These methods further drop the accuracy in low light conditions: 75$\%$, 78$\%$, and 79$\%$ for the EV-SegNet, ESS, and GTNN respectively. The proposed GMNN seems to be robust against varying lighting conditions as it achieves the highest accuracy in both lighting conditions i.e. 95\% in bright and 94\% in dark light.

The subsequent experiment was conducted on a subset of the testing dataset where the robotic arm movement direction was varied as linear, rotational, or partial rotational. In event-based vision sensors, the direction of the robotic arm movement plays a crucial role as perpendicular edges generate more informative event sets compared to parallel edges. 

The impact of the phenomenon is demonstrated in experiment 3 of table  \ref{tab: Segmentation Accuracy in Various conditions}, where EV-SegNet, ESS, and GTNN models have the highest accuracy score for rotational motion 73\%,78\%, 89\% respectively, decreasing in partial rotational to  69\%,74\%, 78\% respectively and the lowest for the linear motion  65\%,68\%, 75\% respectively. In contrast, the proposed GMNN achieves the highest average accuracy score of 93\% in rotational motion, decreasing for  partial rotational motion and linear motion to only 90\% and 84\% respectively.

The camera is placed at the end of the robotic arm thus the camera speed is an important factor while evaluating the robustness of the model. The next experiment was conducted on a subset of the training dataset where the speed of the end effector was varied.  As can be seen in experiment 4 of table \ref{tab: Segmentation Accuracy in Various conditions}, state-of-the-art models have an accuracy of 69\% to 75\% for 0.15 m/s, which drops to 60\% to 71\% for 1m/s. Whereas the proposed GMNN model has the highest accuracy of 93\% for 0.15m/s and it drops to 91\% for 1m/s. The clear impact of the CCM mixing layer in high-speed conditions supports the recovery of the information at contours.

In order to understand the scale invariance of the model an experiment was conducted on a subset of the training dataset where the distance between the platform and the camera was varied to 62 cm and 82 cm. As per the results illustrated in experiment 5 of the Table. \ref{tab: Segmentation Accuracy in Various conditions} there is a minimal impact of the camera and object distance on the accuracy of all the models, and GMNN maintains its superiority.

Table \ref{tab: Quantitative Results} compares the performance of four methods, EV-SegNet\cite{Alonso2019EV-SegNet:Cameras}, ESS\cite{Sun2022ESS:Images}, GTNN\cite{AlkendiNeuromorphicNetwork}, and GMNN, on the Known Object Dataset and the ESD-2 Unknown Object Dataset. The proposed GMNN method achieved the highest mIoU and  accuracy, outperforming all other architectures, while EV-SegNet and ESS had the lowest performance, and GTNN achieved a 74.24\% mIoU, which was still 3.5\% lower than the proposed GMNN model.

However, the performance of all methods dropped when evaluated on the Unknown Object Dataset. The proposed GMNN achieved 89.91\% accuracy, while graph-based methods dropped by 6\% compared to EV-SegNet and ESS. These results justify the use of graph structure for asynchronous events for segmentation challenges as it learns the temporal relationships in a better fashion.

\begin{table}[h!]

   \caption{Quantitative comparison of GMNN against other Asynchronous event fusion methods on the whole ESD dataset. }
   \label{tab: Quantitative Results}
\vspace{-10pt}
   \begin{adjustbox}{width=0.48\textwidth}

    \centering
      {\fontsize{6}{8}\selectfont
    % \begin{tabular}{|c|c|c|c|c|c|c|}
        \begin{tabular}{c cc cc}

  \specialrule{.15em}{.1em}{.1em} 
\multicolumn{1}{c}{\textbf{Methods}} & \multicolumn{2}{c}{Known Obj} & \multicolumn{2}{c}{Unknown Obj} \\ \cline{2-5} 

\multicolumn{1}{c}{}    & \multicolumn{1}{c}{mIoU \%} & \multicolumn{1}{c}{Acc \%} & mIoU \%   & Acc \%           \\
  \specialrule{.15em}{.1em}{.1em} 
    EV-SegNet \cite{Alonso2019EV-SegNet:Cameras} & 7.73 & 76.98  & 5.29 & 53.31   \\
      \specialrule{.05em}{.1em}{.1em}     
    ESS \cite{Sun2022ESS:Images} &  8.92 & 81.59 & 7.01	& 67.29 \\
      \specialrule{.05em}{.1em}{.1em} 
    GTNN \cite{AlkendiNeuromorphicNetwork} & 74.24 & 87.53 & 58.70 & 81.30 \\
      \specialrule{.05em}{.1em}{.1em} 
    GMNN (ours)  & \textbf{78.32} & \textbf{96.91} & \textbf{66.05} & \textbf{89.91} \\
  \specialrule{.15em}{.1em}{.1em} 
   \end{tabular}}
 \end{adjustbox}
    \label{tab: Segmentation On Whole Dataset}
 \end{table}

\subsection{\textbf{Model Size}  }
\label{Parametric analysis}

Table \ref{tab: Parametric Comparision} compares the amount of parameters of GMNN against the other three state-of-the-art methods for panoptic segmentation: EV-SegNet\cite{Alonso2019EV-SegNet:Cameras}, ESS\cite{Sun2022ESS:Images}, and GTNN\cite{AlkendiNeuromorphicNetwork}. GMNN utilizes the least number of parameters at 3.9 million, while the other methods require significantly more, ranging from 5.3 to 22 million. This suggests that GMNN may be more efficient and scalable in terms of model size and training time.

\begin{table}[!hbt]
   \caption{Comparison of model size of GMNN against other Asynchronous event fusion methods }
\vspace{-10pt}
   \begin{adjustbox}{width=0.48\textwidth}

      \fontsize{1.5}{2}\selectfont{
        \begin{tabular}{c c }
  \specialrule{.15em}{.1em}{.1em} 

\textbf{Methods} & \textbf{Parameters}   \\
   \specialrule{.15em}{.1em}{.1em} 

    EV-SegNet \cite{Alonso2019EV-SegNet:Cameras} & 22M \\
   \specialrule{.05em}{.1em}{.1em} 
 
    ESS \cite{Sun2022ESS:Images} & 17M \\
  \specialrule{.05em}{.1em}{.1em} 

    GTNN \cite{AlkendiNeuromorphicNetwork} & 5.3M \\
  \specialrule{.05em}{.1em}{.1em} 

    GMNN (ours) & \textbf{3.9M} \\
  \specialrule{.15em}{.1em}{.1em} 
   \end{tabular}}
 \end{adjustbox}
    \label{tab: Parametric Comparision}
     \setlength{\belowcaptionskip}{-12pt}   

 \end{table}

\subsection{\textbf{Computational Time analysis}}
\label{Computational Time analysis}

Table \ref{tab: computational time} displays the computational time taken by the GMNN and GTNN models, implemented on a Dell desktop and a Google Colab's NVIDIA Tesla K80 GPU. The analysis of 40 event graphs, each lasting 10ms, was conducted in two modes of operation, namely sequential and batch mode. The sequential model in PyTorch processes event graphs successively, while the batch mode processes event graphs as a single batch. The results demonstrate that GMNN outperforms GTNN in terms of computational time in both modes of operation. It should be noted that the graph is constructed using events within 100ms, and its size varies from 1 to a maximum of 10000 events, depending on the number of events triggered within the temporal window. Thus, calculating the standard deviation is essential in understanding the computational time performance. Overall, the proposed approach achieves a significant speed-up of up to one order of magnitude in both batch and sequential modes, which is necessary to handle batches of events concurrently while preserving the high temporal resolution of the sensor.

\begin{table}[h]
\centering
\caption{Comparison of Sequential and Batch Modes}
\vspace{-10pt}

\label{tab:comparison}
   \begin{adjustbox}{width=0.48\textwidth}

\fontsize{7.5}{8}\selectfont{\begin{tabular}{c|c|c}
% \hline

  \specialrule{.15em}{.1em}{.1em} 

\textbf{Model} & \textbf{Sequential-mode} & \textbf{Batch-mode} \\ 
\textbf{ } & $\mu$ $\pm$ $\sigma$ (sec)     & $\mu$ (sec) \\ 
\hline
GTNN & 9.63 $\times$ 10$^{-2}$ $\pm$ 1.93 $\times$ 10$^{-4}$ & 13.52 $\times$ 10$^{-4}$ \\
GMNN & \textbf{4.06 $\times$ 10$^{-3}$ $\pm$ 2.05 $\times$ 10$^{-4}$} & \textbf{10.27 $\times$ 10$^{-5}$} \\
  \specialrule{.15em}{.1em}{.1em} 
\end{tabular}}
\end{adjustbox}
\label{tab: computational time}
\end{table}

\section{Ablation Study}
We conduct an ablation study about the proposed CCM in semantic segmentation on the ESD dataset to understand the best suitable combinations of nearest neighbors and a number of parallel features mixing.

\subsection{\textbf{Parallel Mixing within CCM}}

Table \ref{tab: ablation study - varying no of layers} investigates the effect of varying the number of K-Nearest Neighbours (kNN) in each layer of the CCM on accuracy in deep learning models. Results indicate that an increase in the number of kNN layers generally improves panoptic segmentation accuracy for up to 4 layers, however increasing the number of layers further leads to a decrease. This emphasizes the importance of optimizing the number of layers and kNNs in the CCM for maximum accuracy. Note that the PointMixer \cite{Choe2022PointMixer:Understanding}method is equivalent to using only one kNN layer.

\begin{table}[h!]
   \caption{Parametric comparison of GMNN against other Asynchronous event fusion methods. }
\vspace{-10pt}
   \begin{adjustbox}{width=0.48\textwidth}

    \centering
      {\fontsize{6}{7}\selectfont
    % \begin{tabular}{|c|c|c|c|c|c|c|}

        \begin{tabular}{c|ccccccc|c}

  \specialrule{.15em}{.1em}{.1em}

    %\cline{2-8}
    \multicolumn{1}{c|}{\textbf{}} & \multicolumn{7}{|c|}{\textbf{kNN in Each Layer of CCM}} & \\
  \specialrule{.15em}{.1em}{.1em} 

\textbf{} & \textbf{L1} & \textbf{L2} & \textbf{L3} & \textbf{L4} & \textbf{L5} & \textbf{L6} & \textbf{L7} & \textbf{ACC\%}\ \\
  \specialrule{.05em}{.1em}{.1em} 
    1 & 16 & - & - & - & - & - & - & 81.23 \\
  \specialrule{.05em}{.1em}{.1em} 
    2 & 16 & 32 & - & - & - & - & - & 89.04  \\
  \specialrule{.05em}{.1em}{.1em} 
    3 & 16 & 32 & 48 & - & - & - & - & 92.50 \\
  \specialrule{.05em}{.1em}{.1em} 
    4 & 16 & 32 & 48 & 64 & - & - & - & \textbf{96.91} \\
  \specialrule{.05em}{.1em}{.1em} 
    5 & 16 & 32 & 48 & 64 & 80 & - & - & 96.05 \\
  \specialrule{.05em}{.1em}{.1em} 
    6 & 16 & 32 & 48 & 64 & 80 & 96 & - & 95.37  \\
  \specialrule{.05em}{.1em}{.1em} 
    7 & 16 & 32 & 48 & 64 & 80 & 96 & 112 & 93.75 \\
  \specialrule{.15em}{.1em}{.1em} 
   \end{tabular}}
   
    \end{adjustbox}
    \label{tab: ablation study - varying no of layers}
 \end{table}

\vspace{-1cm}

\subsection{\textbf{Impact of varying the kNN size}}

Table \ref{tab: ablation study - set of K} investigates the effect of different sets of nearest neighbors in the CCM layer on the accuracy. Over-smoothing can occur in CCM when too many features are stacked together. The study found that increasing the set of k initially improves accuracy, but Set 4  and Set 5 resulted in decreased accuracy due to over-smoothing.

\begin{table}[h!]
   \caption{Parametric comparison of GMNN against other Asynchronous event fusion methods. }
\vspace{-10pt}
   \begin{adjustbox}{width=0.48\textwidth}

    \centering
      {\fontsize{10}{12}\selectfont
    % \begin{tabular}{|c|c|c|c|c|c|c|}
       
        \begin{tabular}{cccccc}

  \specialrule{.15em}{.1em}{.1em} 
\textbf{\begin{tabular}{c}Set of\\ k \end{tabular}} & \textbf{\begin{tabular}{c}Layer \\1\end{tabular}} & \textbf{\begin{tabular}{c}Layer\\ 2\end{tabular}} & \textbf{\begin{tabular}{c}Layer\\ 3\end{tabular}} & \textbf{\begin{tabular}{c}Layer\\ 4\end{tabular}} & \textbf{\begin{tabular}{c}Layer\\ 5\end{tabular}}\ \\
  \specialrule{.15em}{.1em}{.1em} 
    Set 1 & 3 & 3 & 9 & 12 & 89.23\% \\
  \specialrule{.05em}{.1em}{.1em} 
    Set 2 & 8 & 16 & 24 & 32 & 93.04\%  \\
  \specialrule{.05em}{.1em}{.1em} 
    Set 3 & 16 & 32 & 48 & 64 & \textbf{96.91}\% \\
  \specialrule{.05em}{.1em}{.1em} 
    Set 4 & 25 & 50 & 75 & 100 & 95.19\% \\
  \specialrule{.05em}{.1em}{.1em} 
    Set 5 & 40 & 80 & 160 & 240 & 92.50\% \\
  \specialrule{.15em}{.1em}{.1em} 
   \end{tabular}}
   \end{adjustbox}
    \label{tab: ablation study - set of K}
 \end{table}

\subsection{\textbf{Misclassified Boundary}}

Table \ref{tab: Event overlap} compares four panoptic segmentation methods for known and unknown object subsets using TP, FP, TN, and FN evaluation metrics. The percentage of misclassified events, representing the percentage of FP events that overlap with object boundaries, was measured using 84000 events for known objects and 32000 events for unknown objects. Segmenting the boundaries of an object is significant for robotic grasping because it allows the robot to precisely locate and segment the object, enabling planning and executing more accurate grasping motions, improving efficiency and reducing the risk of mishandling. The GMNN method had the lowest percentage of the misclassified events on the boundaries of 4\% for known and 10\% for unknown object subsets, while other methods ranged from 10\% to 12\% for known and 20\% to 35\% for unknown objects. These findings suggest that the GMNN method is better at identifying object edges, making it a promising option for robotic grasping in challenging conditions.

\begin{table}[t]
    \centering
    \caption{Analysis of event overlap.}
    \vspace{-10pt}
   \begin{adjustbox}{width=0.48\textwidth}

    \centering
      {\fontsize{10}{12}\selectfont
    % \begin{tabular}{|c|c|c|c|c|c|c|}
       
    \begin{tabular}{lccccc}
  \specialrule{.15em}{.1em}{.1em} 
        Method & TP & FP & TN & FN &\begin{tabular}{c} Overlapped \\ Events \\ (\% of FP)\end{tabular}  \\
  \specialrule{.15em}{.1em}{.1em} 
        \multicolumn{6}{c}{Known Object} \\
  \specialrule{.1em}{.1em}{.1em} 
        EV-SegNet\cite{Alonso2019EV-SegNet:Cameras} & 20160 & 10080 & 41160 & 12600 & 12\%  \\
        ESS\cite{Sun2022ESS:Images}& 25200 & 9240 & 42840 & 6720 & 11\% \\
        GTNN\cite{AlkendiNeuromorphicNetwork} & 36960 & 8400 & 35280 & 3360 & 10\% \\
        GMNN (ours) & 42000 & 3360 & 37800 & 840 & \textbf{4\%} \\
  \specialrule{.1em}{.1em}{.1em} 
        \multicolumn{6}{c}{Unknown Object} \\
  \specialrule{.1em}{.1em}{.1em} 
        EV-SegNet\cite{Alonso2019EV-SegNet:Cameras} & 4902 & 11098 & 14184 & 1816 & 35\% \\
        ESS \cite{Sun2022ESS:Images} & 3778 & 12243 & 15128 & 872 & 38\% \\
        GTNN\cite{AlkendiNeuromorphicNetwork} & 9444 & 6556 & 13580 & 2420 & 20\% \\
        GMNN (ours) & 12806 & 3194 & 13026 & 2974 & \textbf{10}\% \\
  \specialrule{.15em}{.1em}{.1em} 
    \end{tabular}}
    \end{adjustbox}
    \label{tab: Event overlap}
     \setlength{\belowcaptionskip}{-12pt} 
\end{table}

 \vspace{-0.6cm}

\section{Conclusion}

The study proposes an approach to panoptic segmentation using a dynamic vision sensor and integrating the novel Collaborative Contextual Mixing (CCM) technique with the U-Net framework. The architecture combines neighboring events at multiple levels and produces a parallel feature learning representation. The encoder performs downsampling operations while the decoder executes upsampling operations on events, resulting in an effective panoptic segmentation model for robotic grasping.

Our proposed model performs exceptionally well on the ESD dataset under diverse conditions and achieves state-of-the-art results in terms of mIoU and pixel accuracy, demonstrating the robustness of the introduced CCM approach against challenges like occlusions, low lighting, small objects, high speed, and linear motion. Additionally, our method utilizes GNNs and mixer techniques, resulting in shorter prediction time than existing state-of-the-art methods.

Future research could explore the proposed approach's generalization capability in real-world scenarios with diverse robots, sensors, and environments, and incorporating other sensors like depth sensors or thermal cameras that could improve the model's low-light performance.

% \newline
% \textbf{Acknowledgements}
\section*{Acknowledgements}
This academic publication was funded by Kingston University, the Advanced Research and Innovation Center (ARIC), and Khalifa University of Science and Technology, Abu Dhabi, UAE.
%and acknowledges the significant contributions of researchers from Khalifa University. 
In addition, the authors would like to extend their appreciation to Ipsotek, an Eviden Company, for their generous support and valuable contributions to the research project.

\bibliographystyle{IEEEtran}

\bibliography{references.bib}

\end{document}